# Ubiquitous WLAN/Camera Positioning using Inverse Intensity Chromaticity Space-based Feature Detection and Matching: A Preliminary Result

Wan Mohd Yaakob Wan Bejuri[1], *Associate Member, IEEE*, Mohd Murtadha Mohamad[1], *Member, IEEE*, Maimunah Sapri[2] and Mohd Adly Rosly[1]

[1]Faculty of Computer Science & Information Systems,
Universiti Teknologi Malaysia, Johor, Malaysia
wanmohdyaakob@gmail.com, murtadha@utm.my, madlysalaf@gmail.com
[2]Center of Real Estate Studies,
Universiti Teknologi Malaysia, Johor, Malaysia
maimunahsapri@utm.my

*Abstract*—**This paper present our new intensity chromaticity space-based feature detection and matching algorithm.** This approach utilizes hybridization of wireless local area network and camera internal sensor which to receive signal strength from a access point and the same time retrieve interest point information from hallways. This information is combined by model fitting approach in order to find the absolute of user target position. No conventional searching algorithm is required, thus it is expected reducing the computational complexity. Finally we present pre-experimental results to illustrate the performance of the localization system for an indoor environment set-up.

## I. INTRODUCTION

The movement into the ubiquitous computing realm will integrate the advances from both mobile and pervasive computing. Though these terms are often used interchangeably, they are conceptually different and employ different ideas of organizing and managing computing services The increasing of mobile usage has been widespread since it is a most cool gadget that compromise mobility, efficiency and effective to the end users. Today, mobile phone is not just as communicator device, it is also can be navigator device. Most of the high-end mobile device has been equipped with GPS navigation system that gives navigation services to end user if they want to travel to interest destination. However, the navigation by using Global Positioning System (GPS) was suffered in obstruction environment especially user inside building (indoor environment) [1] [2] [3] [4] [5] [6] [7] [8] [9] [10]. In addition, the object such as tree, high building, high wall and also people walking might be the contributors of the obstruction. These obstructions sometimes moved to another location which usually happened in indoor environment and finally make it difficult to estimate user's position. Therefore, there is a need an alternative way in order to ensure users can locate herself or himself inside building too (For example situation: a visitor want to find her friend in a complex building office) The integration of GPS with others positioning sensor may help to solve this issue in order to determine positioning information with more intelligent, reliable and ubiquity. Although the integration of GPS with external sensor such as inertial navigation system (INS) are known quite successful in term of navigation inside building, this solution are not tend very much to solve this issue since it may make end users feel harsh on device integration. The integration with external sensor mostly is quite successful on positioning accuracy, but it is not really successful in terms of mobility. On the other side, the integration GPS with internal positioning sensor such as WLAN or, Camera or Bluetooth may solve this issue.

In this paper, we are focusing the indoor positioning in term to how to get positioning information inside building in a hallways/corridor by using hybridization of WLAN and Camera. The structure of the paper is as follows. Section 2 will present the reviews related work to hybrid WLAN/Camera and others indoor positioning. Section 3 will present an overview of the our proposed method. The details of our preliminary result are covered in section 4. Finally, conclusions are given in section 5.

## II. RELATED WORK

Although we believe our proposed algorithm is novel, previous method can locate their device by using camera sensor in the environments. Most of these techniques are developed for embedded navigation system such as robot by using input data from mechanical based sensor and image. In the early works of research shows that matching building features is feasible [11], but unfortunately, the requirement of measurement such as positioning need to accurate. Most of the recent research uses robust image features technique such as produced by Scale Invariant Feature Transform (SIFT). The technique based on builds map simultaneously and also locate position of interest image features within that map [12].

This research sponsored by Research University Grant (Title: The Prototype of Mobile Ubiquitous Positioning for Unmovable Physical Facility Tracking, Project Vot No: QJ13000.7128.02J56)

However, those techniques are not suitable for our scenario, where external mechanical sensor such inertial navigation system (INS) are not available within our scope.

In the issue that relate with 3D locations of feature points, the Photo Tourism system can be used. This software also can be used to compute positioning information of camera from a large group [13]. This system works by matches new images captured to the current model, in order to establish an accurate camera pose in relation to the scene. It is also can send annotations between images by using the underlying 3D structure. However, this system just support in open areas not in hallways.

Beside of that, Harlan Hile [14] was come out with her new research based on positioning information computation using camera. Current approaches to this problem use location systems to help the user index into the information. The hybridization of WLAN and Camera has been used in order to provide a contextual framework on a mobile phone. The phone's display then becomes the interface for overlaying information directly onto the image and thus providing the user information in context. Although the result shows this method is quiet good, but it is also suffer with illumination change.

Although these techniques also support similar interactions, none of them is support in illumination change environments. The system based on Augmented Reality (AR) also shared a similar objective for information overlay. These kinds of the systems tend to be object centric. They usually tag objects with AR markers in order to estimate their location in image. [15]. The others, is actively project structured light on the environment to aid in location determination [16], but it have some difficulty in the hallway. Most of the existing system based on AR provides a variety of pseudocode (information-overlay) and might be our algorithm source, but they do not currently support our research scope (hallway environment) without using additional special tagging or hardware.

### III. METHODOLOGY

Generally, our concept is to determine positioning by using mobile phone without external sensor. In order to archive positioning precision, the integration between different sensor inputs may help to solve situation in indoor environment. However, the limitation of camera in illumination environment could affect the performance of system such as latency. This illumination noise which is contain together with useful image information in image data will processed together may make latency occurred in matching part. In our approach, we used two type of input which is WLAN and camera. In Figure 1 illustrates how our algorithm will work. The input from camera will extracted in order to obtain feature interest (corner) and the same time, the input from WLAN will extracted in order to gather WLAN positioning coordinate. This type of information will send directly trough wireless network to server. In the server, the image input will firstly processed by illumination algorithm in order to reduce illumination error in image input. Then it will processed by image segmentation method before it processed through corner detection process in order to get feature interest point. After image information processing, this interest point image will matched with coordinate information that obtained from WLAN positioning by using model fitting approach.

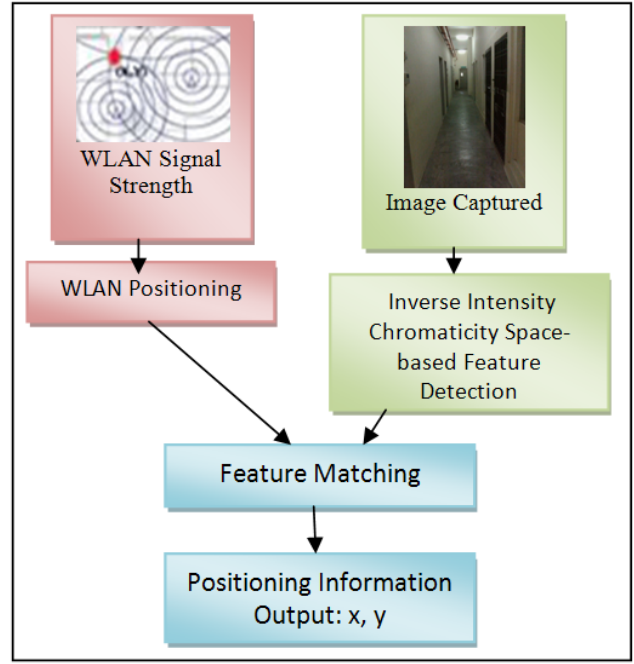

Figure 1. Flow Diagram for Calculating Indoor Positioning Information. There are Two (2) Different Input in Order to Generate The Positioning Output.

#### A. Inverse Intensity Chromaticity Space-based Feature Detection in Hallways/Corridor

In feature detection part, our algorithm will locate image features as a first step. However, there are some limitation of location information in the building map, the map suppose have another information including door and corner. Usually these type information we can calls as micro-landmarks which located (in the image) at the doorways lines and floor edge corners meet. In this part, we developed a feature detection based on inverse intensity chromaticity space in order to obtain such information interest point on the floor edges. However, there are noise will occurred such illumination error, in this type of error, we choosing inverse intensity chromaticity space method in order to reduce this kind of error. The mean shift [17] method is used in order to segment the image. After the images have been segmented, this algorithm will detect interest corner at edge intersection of doorways and floor by using cornerity metric [18]. However, we believe this type of corner detection cannot search all suitable corner, but at least it find many of them. The false detection of corner won't correspond to the floor plan since occlusion of the floor.

*B. Features Matching*

When features interest information has been obtained, the next step is to integrate this information together with WLAN positioning information. The WLAN [19] will give positioning information within 10m accuracy in order to select which possibility hallways that user should be located. Additionally, it is provides estimated gross center for regional to be tested; matching system can also search radius estimate based on accuracy and some ideas useful horizontal visibility location system camera. The integration of two (2) different information will finally make correspondence between image captured and floor plan. The reason behind this is to reduce ambiguous cases and search space positioning information in database.

For the correspondence part, there are too much possible correspondence. For example, let say if we want to match 10 point in a image captured with 32 point in a floor map, the possibility corresponding match between image and floor point may results 4 billion possible four-point correspondences. In order to solve this issue, we prefer to use Random Sample Random Sample Consensus (RANSAC) [20] that operates to select and optimize the hypotheses. In this part, we are using minimal structural assumptions to generate the hypotheses for the RANSAC algorithm in order to fit lines in the image-space features, at which produce left and right correspondence line. Then, the algorithm choose as random two (2) points from each line in image captured that orders the points along the lines consistently. More larger distance between 2 points, means more stable likely produce stable camera.

## IV. PRELIMINARY RESULT

The experimental testbed is located at faculty of Computer Science & Information System, Universiti Teknologi Malaysia building. Consequently, we just implement on single floor although in this work was conducted which our experiments only on locations in the 3th floor of the building. The floor has dimensions of 45 feet by 105 feet and includes more than ten (10) rooms. The data collection was obtained by taking image at five (5) different interest location in which focusing in main hallways. The image was captured by using personal digital assistant (PDA) model HTC HD Mini with supported camera up to 5 Mega Pixels in a day environment only. The image captured that seen by the mobile phone camera can vary depending on the illumination condition on that environment. On the contrary, the lighting source (for example: lamp) is weaker, and the illumination may have a good condition. Below, we will discuss more detail about performance of proposed approach.

Referring to Figure 2, there are eight (8) interest points detected. This location of this image was referred as location A. At the left side in the image, we can see only 1 (one) interest point detected. This is cause by black shadow since poor lighting condition on that area. However, four (4) interest points at in front right side is true corner detected and the others is false interest points since it is suffer to illumination change. The condition of illumination change may make image is not properly segmented.

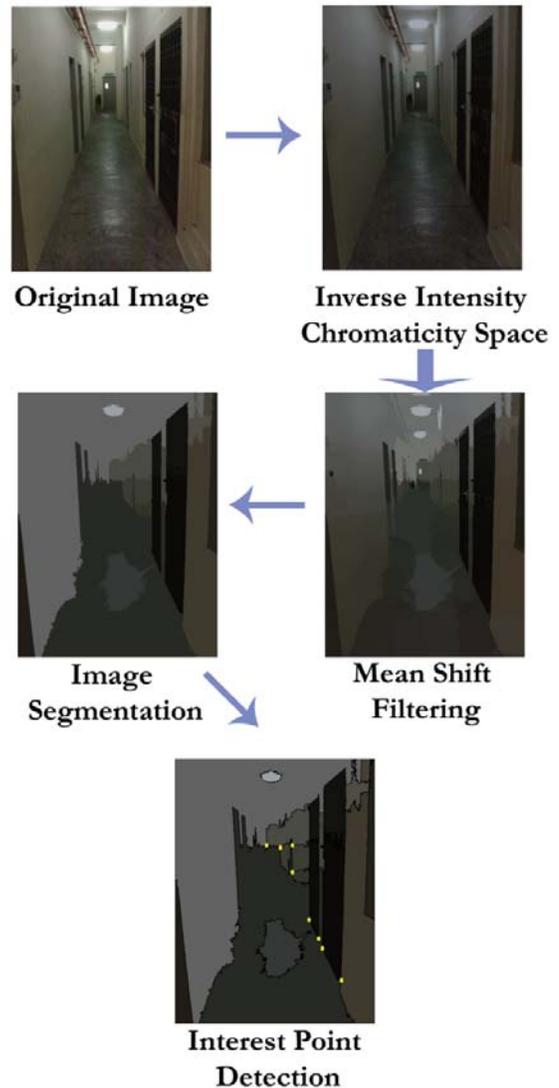

Figure 2: There are eight (8) interest point detected at location A

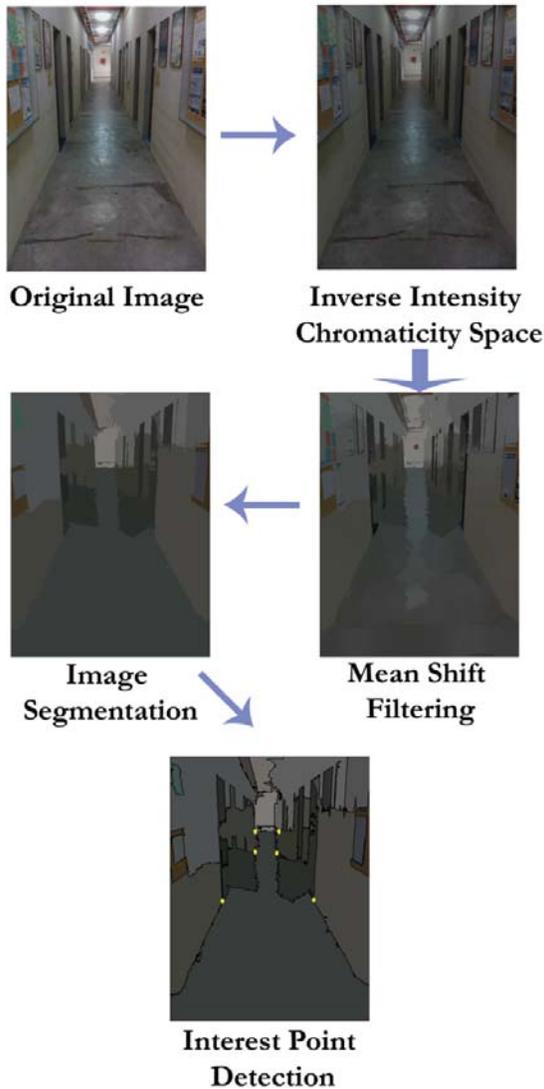

Figure 3: There six (6) interest points detected at location B

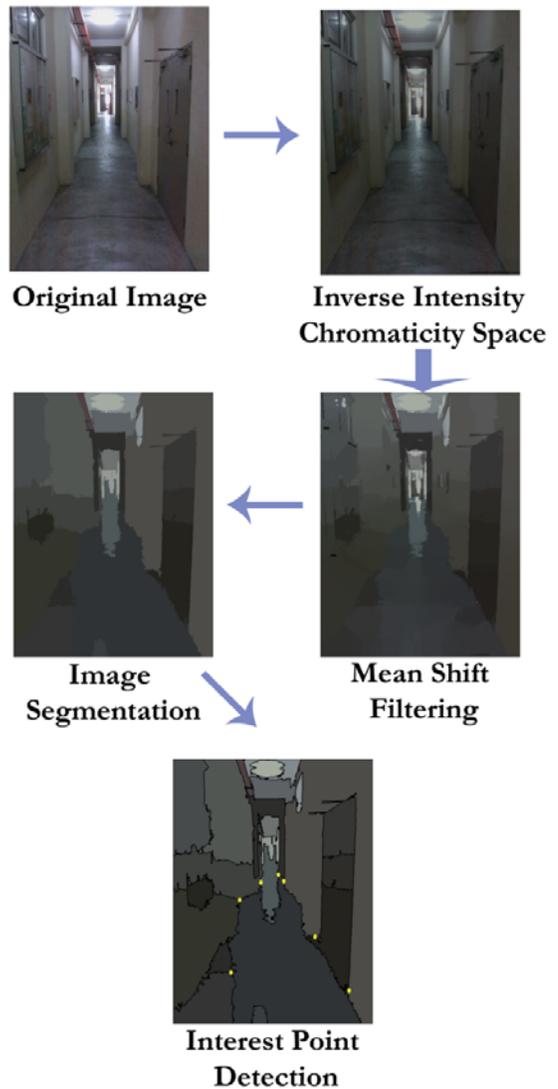

Figure 4: There seven (7) interest points detected at location C

On the Figure 3, there are six (6) interest points detected. This location of this image was referred as location B. Although 4 points (which 2 of them located at in front and others at the back) are detected as true interest points detection, two (2) points at in middle (at left side and right side) are classifies as false interest points detection. It is cause by natural lighting shadow (which is source from sun) founded on hallway floor. It is make the color of a part of hallways floor was affected finally makes image segmentation tool "thinks a part of floor that affected by natural light is only floor and the others is wall". However two (2) of interest points detected at the back not affected by natural light since the position shadow of natural light was exact with position of doorway.

On the Figure 4, there are seven (7) interest points detected. The interest points detection looks like scattered on the hallway floor. This location of this image was referred as location C. Actually, none of the interest points detection are false except two (2) interest points detection on right in front side. This is caused by natural lighting shadow was obstructed by at least two (2) undefined objects at the back. The undefined obstruction may cause the form of natural light shadow is not proper and finally it makes image segmentation process producing a scattered segment result. However, the interest points detection at the right in front side actually near to false although it is can categorize as true.

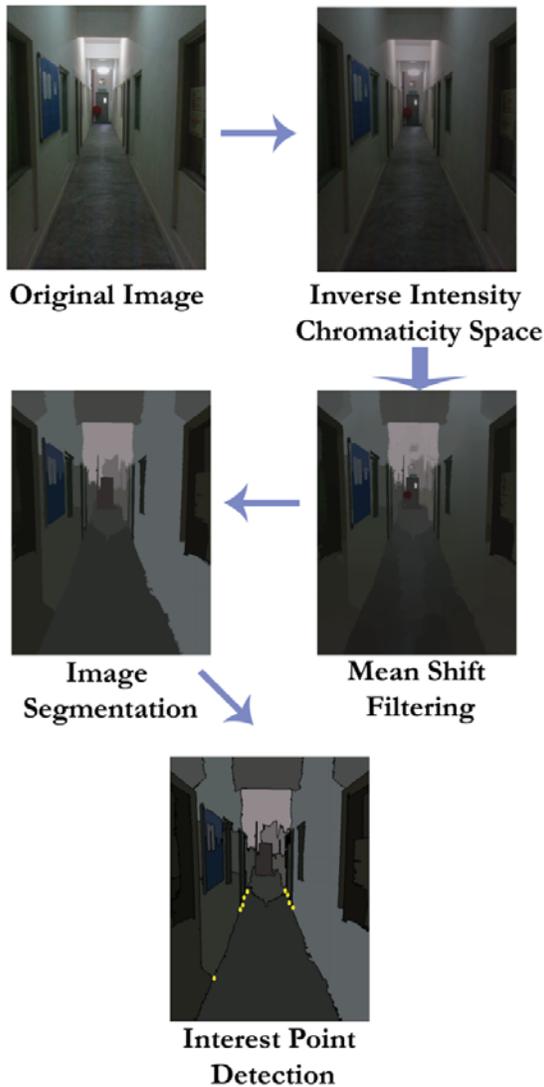

Figure 5: There nine (9) interest points detected at location D

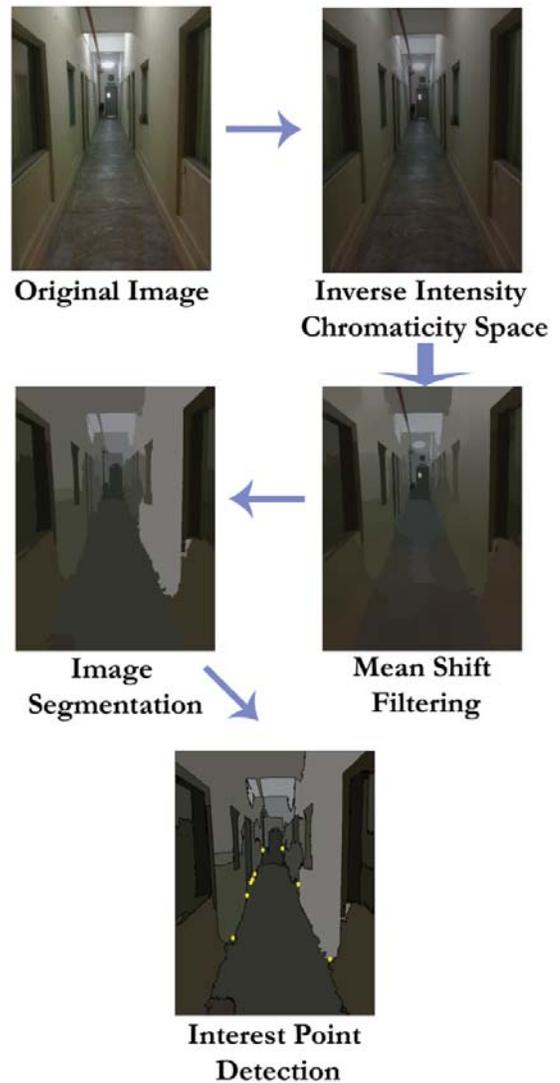

Figure 6: There nine (9) interest points detected at location E

Referring to Figure 5, there are nine (9) interest points detected. This location of this image was referred as location D. At the left side in the image, we can see only one (1) false interest point detected and the others are categorized as true interest point detection since there are founded a notch at building architecture. It is may cause image segmentation process will segment at that are as a region although there is not seen by human visual any door at there.

Referring to Figure 6, there are nine (9) interest points detected. This location of this image was referred as location E. All of these interest detection points are categorized as true except two (2) interest points detection at left in front and in front are categorized as false. It is caused by notch on building architecture at both of side. This case almost same with location D where there notch on building architecture founded. The image segmentation process will accept that region notch as a region although in real situation there is no door founded.

## V. CONCLUSION

This paper discussed about our pre experiment for location determination by using intensity chromaticity space-based feature detection and matching algorithm on hybridization of wireless local area network and camera. The information from both of sensor output was combined by model fitting approach in order to find the absolute of user target position, without using any conventional searching algorithm. The preliminary result shows, most of the experiment were suffered with illumination change. The illumination change will affect position and total of interest points in image. As a future works, we will continue our experiment by using this result and combine with WLAN positioning data in order to know how far our approach can affect user target position in a hallways.

ACKNOWLEDGMENT

This paper was inspired from my master research project which is related to indoor positioning on mobile phone. The author also would like to thank our supervisor Mohd Murtadha b Mohamad and also our co-supervisor Dr. Maimunah bt Sapri for his insightful comments on earlier drafts of this paper.